\documentclass[chi_draft]{sigkddExp}
\usepackage{amsmath,amssymb}
\usepackage[textsize=tiny]{todonotes}
\usepackage{marginnote}\let\marginpar\marginnote
\usepackage[pdfborder={0 0 0}]{hyperref}
\usepackage{url}
\usepackage{csquotes}
\usepackage{subcaption}

\newcommand{\SSE}{\ensuremath{\operatorname{SSE}}}
\newcommand{\Var}{\ensuremath{\operatorname{Var}}}

\begin{document}
\title{
Stop using the elbow criterion for k-means
}%
\subtitle{
and how to choose the number of clusters instead
}

\numberofauthors{1}
\author{
\alignauthor
Erich Schubert\\
\affaddr{TU Dortmund University}\\
\affaddr{44221~Dortmund, Germany}\\
\email{erich.schubert@tu-dortmund.de}
}
\date{11 November 2022}

\maketitle

\abstract{
A major challenge when using k-means clustering often is
how to choose the parameter k, the number of clusters.
In this letter, we want to point out that it is very easy to
draw poor conclusions from a common %
heuristic,
the ``elbow method''.
Better alternatives have been known in literature for a long time,
and we want to draw attention to some of these easy to use options,
that often perform better.
This letter is a call to stop using the elbow method altogether,
because it severely lacks theoretic support, and we want to
encourage educators to discuss the problems of the method
-- if introducing it in class at all --
and teach alternatives instead, while researchers and reviewers
should reject conclusions drawn from the elbow method.
}

\section{Introduction}

Cluster analysis aims at identifying subgroups in the data that
have high similarity within the group, while they also differ
from the remainder of the data set.
No single ``best'' definition of a cluster exists.
Bonner~\cite{DBLP:journals/ibmrd/Bonner64} noted that
\enquote{none of the many specific definitions [of clusters]
seems \enquote{best} in any general sense},
and Estivill-Castro~\cite{DBLP:journals/sigkdd/Estivill-Castro02}
argued that it cannot exist.
each data set and use case may call for different properties to be
desirable, which in turn leads to different algorithms to find the ``best'' solution.
Hence, a large number of clustering methods were developed over the
last decades, based on concepts such as finding a hierarchical structure
(akin to phylogenetic trees), quantization and compression,
parametric modeling, or identifying dense areas.

Despite the many different concepts of clusters and the wide variety of
clustering algorithms available, one method currently is the
most used and most taught clustering method: k-means clustering.
One of the main reasons may be the simplicity of the standard algorithm:
assigning each point to the nearest center,
then recomputing all the cluster centers until nothing changes
-- this algorithm can be easily described in a single sentence.
At the same time, this algorithm runs very fast, and it will always produce
a result with exactly $k$~clusters, giving a (false) suggestion of success.
A key problem then with applying this method to data is often the need
to choose the number of clusters~$k$, although users should first consider
whether $k$-means is even the right choice for their problem at all,
and pay more attention to data preprocessing, too. It makes no
sense to search for the ``optimum'' $k$ if $k$-means is not solving
the problem. 

\section{k-means clustering}

Formally, $k$-means clustering is a least-squares optimization problem.
We can best view it as a data quantization technique, where we want to
approximate the data set of $N$ objects in a continuous,
$d$-dimensional vector space
$\mathbb{R}^d$ using $k$ centers.
The quantization error for a data set $X$ and a set $C$ of centers then is
called inertia, the within-cluster sum of squares (WCSS),
or the sum of squared errors (SSE):
\begin{align}
\SSE(X,C) =& \textstyle\sum_{x\in X} \min_{c\in C} \, \lVert x-c \rVert^2
\;.
\label{eq:sse}
\end{align}
While it is easy to optimize this for a single cluster center by taking the arithmetic
average in each dimension, i.e., the data set centroid, the problem is NP-hard
for multiple clusters and higher dimensionality~\cite{DBLP:conf/walcom/MahajanNV09,DBLP:journals/ml/AloiseDHP09}.

Several methods to optimize this objective exist, but because of the hardness,
most heuristics will only find a local fixpoint.
Because of the very common least-squares objective, the standard algorithm
has likely been invented several times independently, as discussed in the
overview of Bock~\cite{Bock2007}.
The standard heuristic for $k$-means is an alternating optimization,
which first assigns each point to the nearest current cluster center,
then updates each cluster center position with the centroid of the points
assigned to it. If we keep assignments unchanged whenever distances are
identical, the algorithm will eventually not find any changes and stop
(because both steps may never worsen the objective function, and there exists
only a finite number of possible cluster assignments).
The standard algorithm has a complexity of $O(Nkdi)$, where $i$ is the number of
iterations (which theoretically could be very high, but usually is small in practice).
This makes it one of the fastest clustering methods we have available, compared to
$O(N^3+N^2d)$ for the standard algorithm for hierarchical clustering, or
$O(N^2d)$ for DBSCAN without index acceleration.
Many improvements have been proposed that avoid repeated computations in the
standard algorithm, nevertheless, this very basic form has become quite popular
again with the rise of parallel processing and GPUs: it is embarrassingly parallel,
and hence very easy to implement both in clusters as well as GPUs.
For this letter, it does not matter which variant of the algorithm we use.

\section{The elbow criterion}

\begin{figure*}
\begin{subfigure}[t]{.19\linewidth}\centering\vskip0pt
\includegraphics[width=\linewidth]{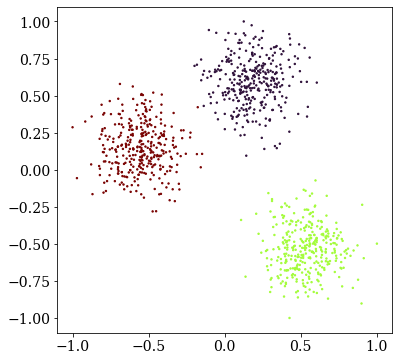}
\caption{Well-separated}
\label{fig:toy-easy}
\end{subfigure}%
\hfill%
\begin{subfigure}[t]{.19\linewidth}\centering\vskip0pt
\includegraphics[width=\linewidth]{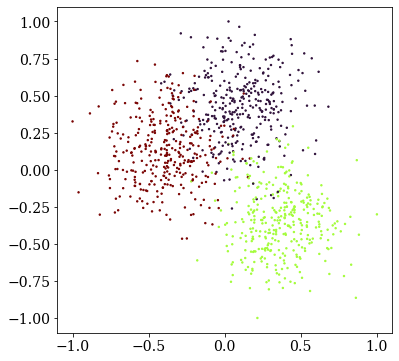}
\caption{Overlapping blobs}
\end{subfigure}%
\hfill%
\begin{subfigure}[t]{.19\linewidth}\centering\vskip0pt
\includegraphics[width=\linewidth]{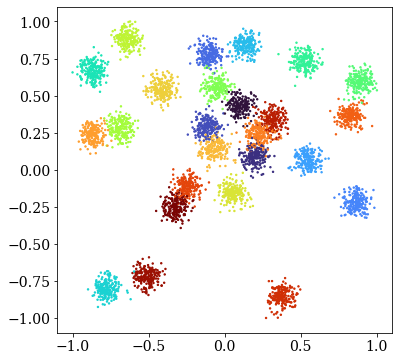}
\caption{Many blobs}
\end{subfigure}%
\hfill%
\begin{subfigure}[t]{.19\linewidth}\centering\vskip0pt
\includegraphics[width=\linewidth]{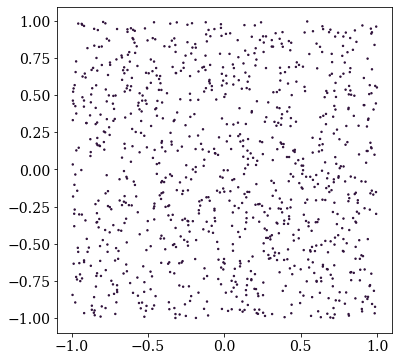}
\caption{Uniform noise}
\end{subfigure}
\hfill%
\begin{subfigure}[t]{.19\linewidth}\centering\vskip0pt
\includegraphics[width=\linewidth]{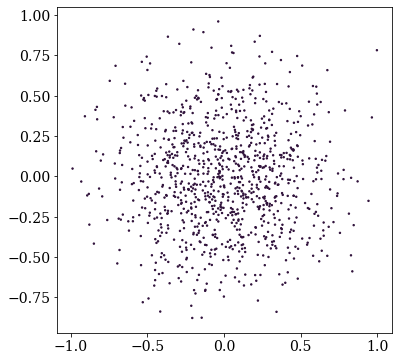}
\caption{Normal noise}
\end{subfigure}
\\
\begin{subfigure}[t]{.19\linewidth}\centering\vskip0pt
\includegraphics[width=\linewidth]{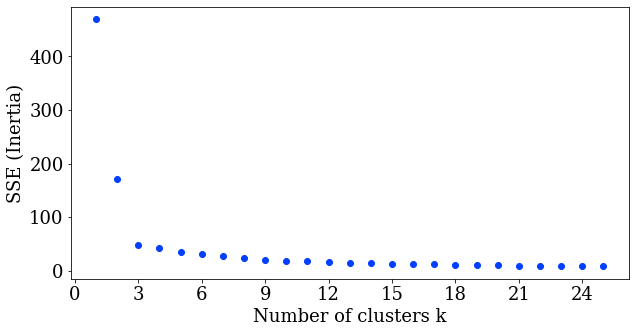}
\caption{Well-separated, $k\leq 25$\vphantom{j}}
\label{fig:toy-easy-elbow}
\end{subfigure}%
\hfill%
\begin{subfigure}[t]{.19\linewidth}\centering\vskip0pt
\includegraphics[width=\linewidth]{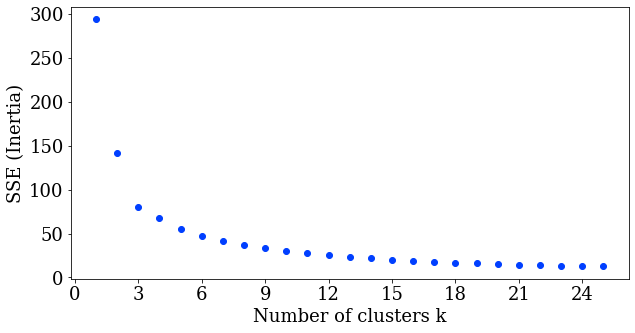}
\caption{Overlapping, $k\leq 25$\vphantom{j}}
\end{subfigure}%
\hfill%
\begin{subfigure}[t]{.19\linewidth}\centering\vskip0pt
\includegraphics[width=\linewidth]{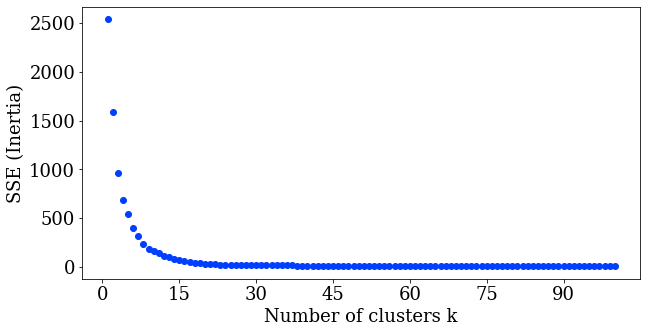}
\caption{Many blobs, $k\leq 100$\vphantom{j}}
\end{subfigure}%
\hfill%
\begin{subfigure}[t]{.19\linewidth}\centering\vskip0pt
\includegraphics[width=\linewidth]{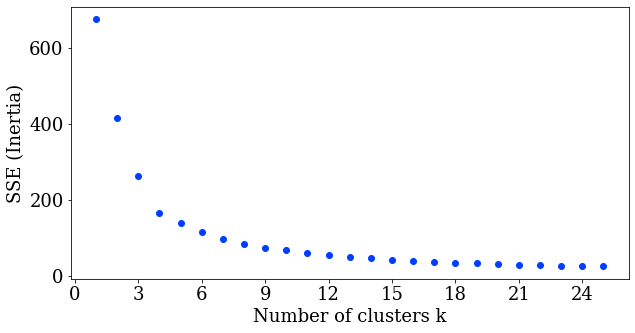}
\caption{Uniform noise, $k\leq 25$\vphantom{j}}
\end{subfigure}
\hfill%
\begin{subfigure}[t]{.19\linewidth}\centering\vskip0pt
\includegraphics[width=\linewidth]{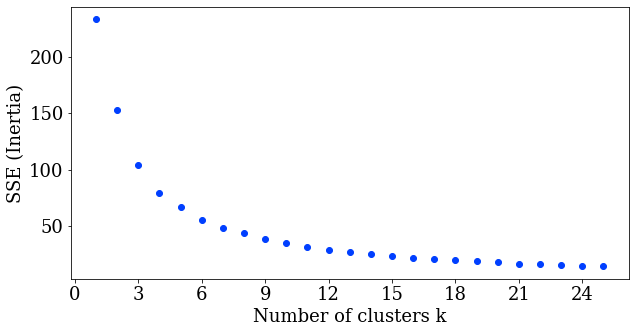}
\caption{Normal noise, $k\leq 25$\vphantom{j}}
\end{subfigure}
\caption{Toy data sets and resulting -- very similar -- elbow plots.}
\label{fig:toy}
\end{figure*}

When the number of clusters~$k$ is not already given by the application,
we have to choose this value; and it turns out this can be rather tricky.
The elbow plot is a chart plotting the approximation error $\SSE$ on the $y$-axis
over a range of values for $k$ on the $x$-axis.
The motivation of the elbow criterion is the concept of diminishing returns:
as we increase the number of clusters, the approximation error decreases.%
\footnote{
Given a $k$-means solution for some $k$, we can trivially construct a
solution with $k+1$ that is better (unless the error already is zero) by
simply adding any point with non-zero error as an additional center.
}
In a data set with very well-separated clusters, we expect to
initially see a sharp drop until some ``optimum'' number of clusters,
afterwards we are splitting ``true'' clusters, which leads to much smaller gains.
And indeed, on certain toy data sets, this appears to work well.
In Figure~\ref{fig:toy-easy} we have a data set with three well-separated clusters,
that is easily clustered by $k$-means. Figure~\ref{fig:toy-easy-elbow} is the
corresponding elbow plot, with a clear inflection at the desired $k=3$.
But the other examples in Figure~\ref{fig:toy} show that the plot always
looks similar, even on uniform data or when the data contains a single normal distribution.

The elbow method is attributed to Thorndike~\cite{Thorndike53},
albeit he notes
\enquote{The curves do not provide much support for the intuitive specification
of the number of clusters}, and concludes with doubt:
\enquote{At this point I can sense the bubbling up of doubts and questions:
\enquote{But what about your \emph{units}?}}.

There are several problems associated with the elbow plot, that statisticians
know too well from the scree plot.
Because the axes of the plot have very different meanings, we cannot compare
them well. We do not have a meaningful measurement of angle, and changing the
scaling of the axes (and, e.g., the parameter range of $k$) may well change the
human interpretation of an ``elbow''.

\subsection{Elbow Detection}
Several attempts to formalize the notion of an ``elbow'' can be found in
software and literature. We present only an excerpt in the following,
largely to illustrate how \emph{heuristic} and \emph{visual} the
machine learning community currently approaches this problem,
instead of improving theory.

Sugar et al.{}\cite{journals/jasa/SugarJ03} propose a ``jump method'',
finding the maximum of $\SSE_k^{-Y}-\SSE_{k-1}^{-Y}$ where
$Y$ is a power parameter suggested to be half the dimensionality. 

Salvador et al.{}~\cite{DBLP:conf/ictai/SalvadorC04} propose the L-method,
which fits linear functions to the points before and after the break;
choosing the breaking point where this piecewise-linear approximation fits
best.
But as discussed above, we will often see an exponential curve, and such a
linear approximation often does not fit this curve at all. To improve this,
the authors also suggest an iterative approach where they truncate the
plot to the first $2\cdot k$ values if $k$ is the best solution found. %

Satopää et al.{}~\cite{DBLP:conf/icdcsw/SatopaaAIR11} in their Kneedle
algorithm want to measure the curvature. For this they fit a smoothing
spline to the data, normalize it to 0 to 1, and compute the difference to the
diagonal. The last maximum before a parameterizable stopping threshold is chosen.

Zhang et al.{}\cite{DBLP:journals/isci/ZhangMQG17} note that the standard
curvature definition is not independent of rescaling the data,
and propose to choose the maximum of a modified curvature:
$$
\text{Curvature}_k :=
\frac{\SSE_{k-1}-\SSE_k}{\SSE_{k}-\SSE_{k+1}} - 1
$$

The pyclustering library \cite{Novikov2019} defines an elbow length:
$$
\text{ElbowLen}_k := \frac{\left(y_0{-}y_1 \right)x_k + \left(x_1{-}x_0\right)y_k + \left( x_0y_1-x_1y_0 \right)}%
{\sqrt{\left( x_1{-}x_0 \right)^2 + \left(y_1{-}y_0\right)^2}}
$$
where $x_0,x_1,y_0,y_1$ denote the minima and maxima of the graph; intended to measure
the length when approximating the curve with the elbow point.
There is no literature given for this approach, the given reference to Thorndike~\cite{Thorndike53}
does not entail this equation, which again appears to depend very much on the scaling of the plot.

Shi et al.{}~\cite{DBLP:journals/ejwcn/ShiWWWLL21} note that
\enquote{experienced analysts cannot clearly identify the elbow point},
and suggest applying a min-max scaling to the range 0 to 10 instead,
then computing angles between triples of adjacent values. The performance
of this approach depends much on this weighting factor.

Onumanyi et al.{}~\cite{Onumanyi22} propose AutoElbow, for which they
min-max scale the elbow plot to 0 to 1, then propose a geometrically-motivated
measure of how close a point is to the bottom left corner and bottom line.
The result of this objective changes substantially when increasing the candidate range of $k$,
and hence likely should not be used at all.
$$\text{AutoElbow}_k := \frac{(x_k-1)^2+(y_k-1)^2}{x_k^2+2\cdot y_k^2}$$

\begin{table}[tb]
\caption{``Optimum'' $k$ chosen by different heuristics on the toy data sets.
$\dagger$~indicates the result can still be recognized as a poor result by the score value or by visual inspection.
$\ddagger$~indicates results that fluctuate with random seeds.}
\label{tab:heuristics}
\setlength{\tabcolsep}{2.5pt}
\begin{tabular}{@{}lc@{\kern9pt}cc@{\kern9pt}cc@{\kern9pt}cc@{\kern9pt}cc@{\kern9pt}cc@{}}
& & \multicolumn{2}{c@{\kern9pt}}{\rotatebox{90}{well-sep.}}
& \multicolumn{2}{c@{\kern9pt}}{\rotatebox{90}{overlapping}}
& \multicolumn{2}{c@{\kern9pt}}{\rotatebox{90}{many blobs}}
& \multicolumn{2}{c@{\kern9pt}}{\rotatebox{90}{uniform}}
& \multicolumn{2}{c}{\rotatebox{90}{normal}}
\\\hline
true $k$ & & \multicolumn{2}{c@{\kern9pt}}{3} & \multicolumn{2}{c@{\kern9pt}}{3} & \multicolumn{2}{c@{\kern9pt}}{25} & \multicolumn{2}{c@{\kern9pt}}{1} & \multicolumn{2}{c}{1} \\
max $k$ & & 10 & 25 & 10 & 25 & 50 & 100 & 10 & 25 & 10 & 25
\\\hline
\multicolumn{12}{@{}c@{}}{\bf Elbow-based}\\
Jump & \cite{journals/jasa/SugarJ03}
& 3 & 3 & 3 & 3 & 23 & 23 & 4 & 4 & 6 & 21 \\
L-Method & \cite{DBLP:conf/ictai/SalvadorC04}
& 3 & 3 & 3 & 4 & 7 & 9 & 4 & 5 & 4 & 5 \\
\multicolumn{2}{@{}l}{L-Method (iter.)} %
& - & 3 & - & 4 & - & 6 & 4 & 4 & 4 & 5 \\
Kneedle & \cite{DBLP:conf/icdcsw/SatopaaAIR11}
& 3 & 3 & 3 & 5 & 8 & 10 & 4 & 5 & 4 & 6 \\
Curvature & \cite{DBLP:journals/isci/ZhangMQG17}
& 3 & 3 & 3 & 3 & 38 & 38 & 4 & 4 & 3 & 21 \\
Pyclustering & \cite{Novikov2019}
& 3 & 3 & 3 & 5 & 8 & 10 & 4 & 5 & 4 & 6 \\
Shi angles & \cite{DBLP:journals/ejwcn/ShiWWWLL21}
& 3 & 3 & 3 & 3 & 3 & 3 & 4 & 4 & 4 & 3 \\
AutoElbow & \cite{Onumanyi22}
& 3 & 3 & 3 & 6 & 9 & 11 & 4 & 6 & 4 & 7 \\
\hline
\multicolumn{12}{@{}c@{}}{\bf Variance-based}\\
Marriot & \cite{journals/biometrics/Marriott71}
& 3 & 3 & 3 & 3 & 25 & 25 & 9\rlap{$^\dagger$} & 17\rlap{$^\dagger$} & 2\rlap{$^\dagger$} & 2\rlap{$^\dagger$} \\
VRC & \cite{doi:10.1080/03610927408827101}
& 3 & 3 & 3 & 3 & 25 & 25 & 9\rlap{$^\dagger$} & 20\rlap{$^\dagger$} & 4\rlap{$^\dagger$} & 4\rlap{$^\dagger$} \\
K-L-Index & \cite{journal/biometric/KrzanowskiL88}
& 7 & 7 & 4 & 10 & 35 & 62 & 5 & 21 & 8 & 8 \\
Pham & \cite{PhamDN05}
& 3 & 3 & 2 & 2 & 8 & 8 & 4 & 4 & 10 & 21 \\
\hline
\multicolumn{2}{@{}l}{Max reduction}
& 3 & 3 & 3 & 3 & 8 & 8 & 4\rlap{$^\dagger$} & 12\rlap{$^\dagger$} & 1\rlap{$^\dagger$} & 1\rlap{$^\dagger$} \\
\multicolumn{2}{@{}l}{Last reduction}
& 3 & 3 & 3 & 3 & 25 & 25 & 9\rlap{$^\dagger$} & 20\rlap{$^\dagger$} & 1\rlap{$^\dagger$} & 1\rlap{$^\dagger$} \\
\hline
\multicolumn{12}{@{}c@{}}{\bf Information-theory-based}\\
BIC & \cite{DBLP:conf/icml/PellegM00}
& 3 & 3 & 3 & 3 & 25 & 100 & 9 & 25 & 1 & 1 \\
BIC (fixed) & \cite{web/FogliaH12}
& 3 & 3 & 3 & 3 & 25 & 25 & 1\rlap{$^\dagger$} & 1\rlap{$^\dagger$} & 1\rlap{$^\dagger$} & 1\rlap{$^\dagger$} \\
\hline
\multicolumn{12}{@{}c@{}}{\bf Distance-based}\\
Dunn & \cite{journal/cybernetics/Dunn73}
& 3 & 3 & 8 & 17 & 18 & 18 & 7 & 20 & 3 & 24 \\
DB & \cite{journals/tpami/DaviesB79}
& 3 & 3 & 3 & 3 & 21 & 21 & 4 & 4 & 10 & 22 \\
Silhouette & \cite{Rousseeuw/87a}
& 3 & 3 & 3 & 3 & 21 & 21 & 4 & 4 & 3 & 3 \\
\multicolumn{2}{@{}l}{Simpl. Silhouette}
& 3 & 3 & 2 & 2 & 21 & 21 & 4 & 4 & 3 & 3 \\
\hline
\multicolumn{12}{@{}c@{}}{\bf Simulation-based}\\
Gap & \cite{journals/jrss/TibshiraniWH01}
& 3 & 3 & 3 & 3 & 30\rlap{$^\ddagger$} & 30\rlap{$^\ddagger$} & 14\rlap{$^\ddagger$} & 21\rlap{$^\ddagger$} & 1 & 1 \\
\end{tabular}
\end{table}

\pagebreak
\subsection{Detection performance}
Several of these measures are sensitive to the range of $k$ that we analyze,
even if the additional values of $k$ perform poorly. They are heuristics based on the geometric
idea of an elbow point, but not taking the process causing the measured data into account.
In Table~\ref{tab:heuristics} we give the results obtained for the toy data sets of
Figure~\ref{fig:toy} using several heuristics proposed in the literature.
Because some methods are very sensitive to the range of $k$ included, we tested two limits,
one rather conservative, and one that is much larger. We observe that all methods
were able to recognize the best solution on the well-separated data set, and even on the
more overlapping version, they all worked -- for a small enough maximum~$k$.
For the data set with many clusters as well as the uniform data set,
all the elbow-based methods failed. The methods based on variance -- which we will discuss
below -- worked much better, but when applied na\"ively will still cluster the uniform data.
Only when using additional thresholds (or visually inspecting the score plot),
the uniform data is recognizable as not clustered. For the normal data, our method
indicates a single cluster, while the classic variance-ratio criterion also discussed
below has a maximum at six clusters.

First of all, the quantity measured, the sum of squared deviations, is a \emph{squared} value.
It would make much more sense to analyze the square root of this value, and if we also take
the number of points into account, the root-mean-squared-deviation (RMSD), which corresponds
to a standard deviation of each point to the nearest center. How meaningful are ``angles'',
``distances'', ``elbows'', and ``slopes'' on a graph that compares $k$ to $\SSE$, two quantities
of different scales?
If we scale the entire data set by a factor of $\alpha$, the $\SSE$ will change by $\alpha^2$,
and the ``optimum'' found by most of the geometric methods changes, while it is clear
that it should~not.

Secondly, increasing the parameter range of $k$ analyzed must not change the decision
once the optimum $k$ is included. Normalizing to the observed minimum and maximum values
(often even starting with $k=2$, not $k=1$) seems inappropriate.
In particular, even without running the algorithm, we know that for $k=N$ we will be able
to get an approximation error of $0$, so we likely should always consider $N$ to be the maximum
$x$ coordinate, and $0$ to be the minimum $y$ coordinate.
A meaningful normalization should preserve~$0$.

Third, we know that even on random data we obtain a descending curve,
and hence we should try to remove this expected behavior from our measure.

Fourth, the method should be able to choose $k=1$ for data that
does not contain any meaningful clusters.

\subsection{Expected behavior of SSE}

Instead of proposing heuristic visual approaches to formalize an imaginary ``elbow'',
we need to first better understand the quantity that we are working with.
The sum of squared errors closely resembles the variance of the data set.
Because our cluster centers are derived from the data, we should be using a
form of sample variance. Simply diving the $\SSE$ by $N$ will be a biased
estimate, and we postulate that $\SSE/(N-k)$ is a more suitable estimate
in this context. But since usually $k\ll N$, this will not make much of a difference yet.
Instead of working with the squared quantity, we then may want to apply the
square root instead, i.e., use $\sqrt{\SSE/(N-k)}$ to have the intuition of
a \emph{standard deviation from the nearest center}.
Still, the plot obtained this way will look similar to what we started with,
and because the square root is a monotone function on the outside, it will not
affect the ordering of results -- it only serves to make the quantity more
interpretable, because ideally, the domain expert should judge whether this is
sufficiently small.

As a baseline ``expected'' behavior, we will for simplicity assume the input
data to be uniformly distributed in a single dimension, but with the variance
of the input data set. The variance of a uniform interval of length $b$ is
$\Var([0;b])=\tfrac1{12}b^2$. If we slice this into $k$ slices of equal
length, each of these has $\Var([0;b/k])=\tfrac1{12}b^2/k^2$, and we obtain
for the resulting total variance $k\cdot \Var([0;b/k])=\tfrac1k \Var([0;b])$.
Because of this observation, we propose to use the na\"ive estimate
$\SSE_1/k$ as normalization factor.
But Krzanowski and Lai~\cite{journal/biometric/KrzanowskiL88} suggest that
$\SSE/k^{\frac2d}$ may be more appropriate than our na\"ive estimate.
We should further include the $N-k$ factor
discussed above. If there is more than one good parameter $k$
(e.g., because there are substructures in the data),
we may also want to compare the solution with the best found so far, e.g., using:
\begin{align}
\widehat{\SSE}_k := \tfrac{N-k}{k} \min_{j=1\ldots k-1} \tfrac{j}{N-j} \SSE_j
\label{eq:norm}
\end{align}

\begin{figure*}[tb]\centering
\begin{subfigure}[t]{.33\linewidth}\centering\vskip0pt
\includegraphics[width=\linewidth]{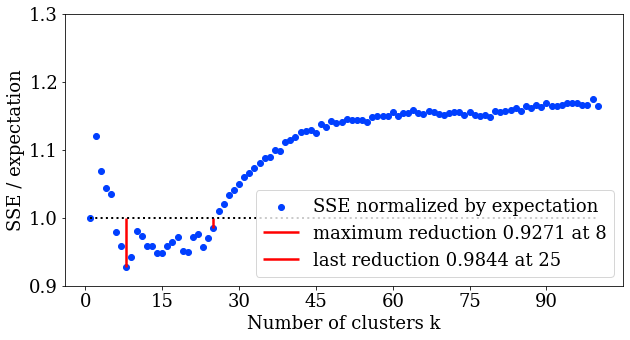}
\caption{Many blobs}
\label{fig:many-reduction}
\end{subfigure}%
\hfill%
\begin{subfigure}[t]{.33\linewidth}\centering\vskip0pt
\includegraphics[width=\linewidth]{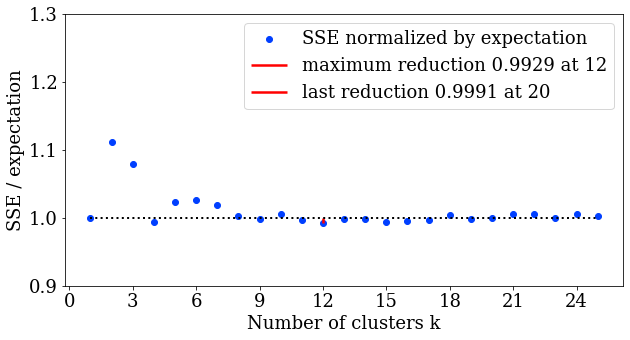}
\caption{Uniform distribution}
\label{fig:unif-reduction}
\end{subfigure}%
\hfill%
\begin{subfigure}[t]{.33\linewidth}\centering\vskip0pt
\includegraphics[width=\linewidth]{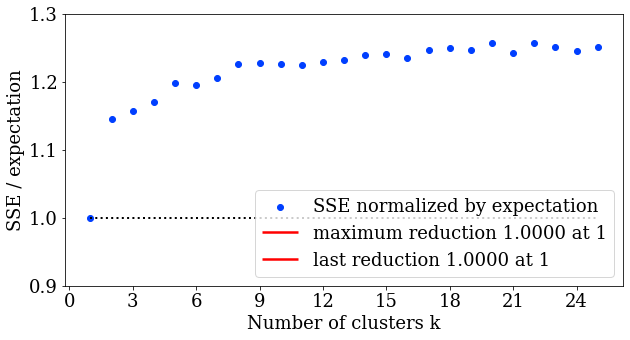}
\caption{Normal distribution}
\label{fig:norm-recution}
\end{subfigure}%
\caption{Reduction in $\smash{\sqrt{SSE}}$ over the estimate $\smash{\sqrt{\widehat{\SSE}_k}}$.}
\label{fig:reduction}
\end{figure*}
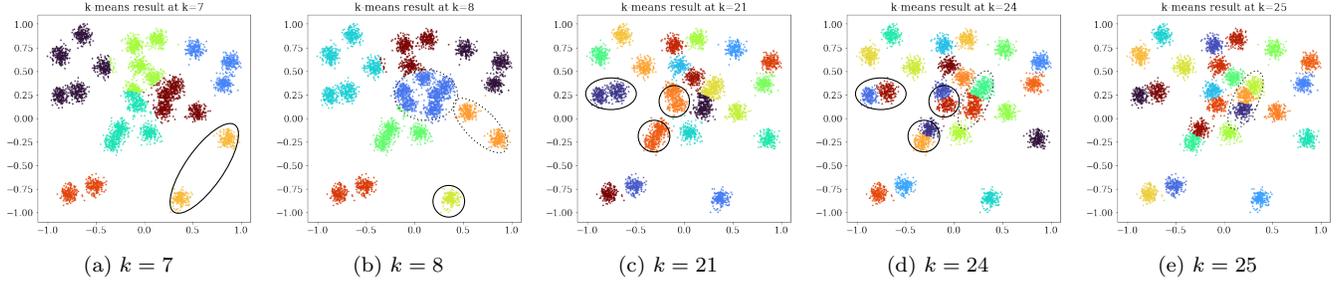
\begin{figure*}[tb]\centering%
\begin{subfigure}[t]{.19\linewidth}\centering\vskip0pt
\begin{tikzpicture}[x=.5\linewidth,y=.5\linewidth,every node/.style={inner sep=0pt}]
\node at (0,0) {\includegraphics[width=\linewidth]{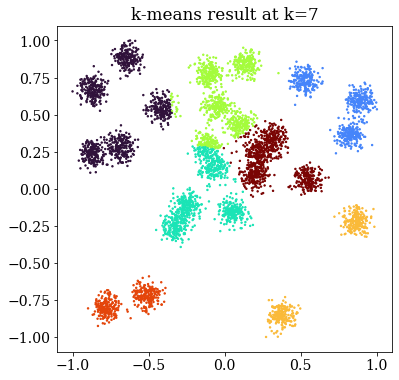}};
\draw[rotate around={-35:(.6,-.39)}] (.6,-.39) ellipse (.15 and .42);
\end{tikzpicture}
\caption{$k=7$}
\label{fig:many-k7}
\end{subfigure}%
\hfill%
\begin{subfigure}[t]{.19\linewidth}\centering\vskip0pt
\begin{tikzpicture}[x=.5\linewidth,y=.5\linewidth,every node/.style={inner sep=0pt}]
\node at (0,0) {\includegraphics[width=\linewidth]{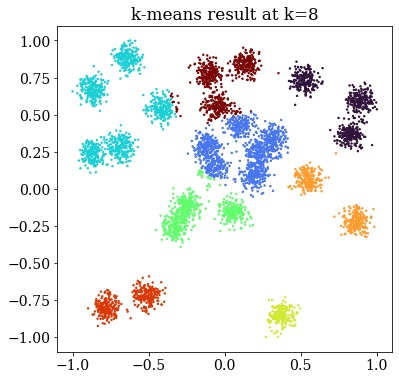}};
\draw[densely dotted] (.22,.19) ellipse (.25 and .2);
\draw[densely dotted,rotate around={45:(.65,-.05)}] (.65,-.05) ellipse (.12 and .28);
\draw (.4,-.65) ellipse (.125 and .125);
\end{tikzpicture}
\caption{$k=8$}
\label{fig:many-k8}
\end{subfigure}%
\hfill%
\begin{subfigure}[t]{.19\linewidth}\centering\vskip0pt
\begin{tikzpicture}[x=.5\linewidth,y=.5\linewidth,every node/.style={inner sep=0pt}]
\node at (0,0) {\includegraphics[width=\linewidth]{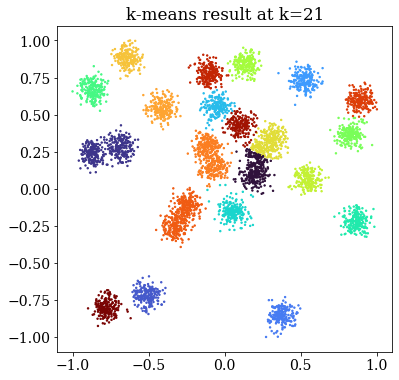}};
\draw (-.45,.2) ellipse (.2 and .125);
\draw (-.11,-.135) ellipse (.125 and .125);
\draw (.05,.14) ellipse (.12 and .125);
\end{tikzpicture}
\caption{$k=21$}
\label{fig:many-k21}
\end{subfigure}%
\hfill%
\begin{subfigure}[t]{.19\linewidth}\centering\vskip0pt
\begin{tikzpicture}[x=.5\linewidth,y=.5\linewidth,every node/.style={inner sep=0pt}]
\node at (0,0) {\includegraphics[width=\linewidth]{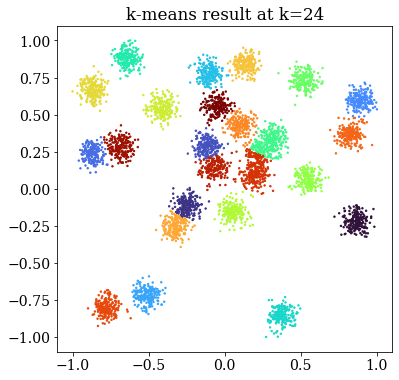}};
\draw (-.45,.2) ellipse (.2 and .125);
\draw (-.11,-.135) ellipse (.125 and .125);
\draw (.05,.14) ellipse (.12 and .125);
\draw[densely dotted,rotate around={-25:(.3,.15)}] (.3,.15) ellipse (.1 and .25);
\end{tikzpicture}
\caption{$k=24$}
\label{fig:many-k24}
\end{subfigure}%
\hfill%
\begin{subfigure}[t]{.19\linewidth}\centering\vskip0pt
\begin{tikzpicture}[x=.5\linewidth,y=.5\linewidth,every node/.style={inner sep=0pt}]
\node at (0,0) {\includegraphics[width=\linewidth]{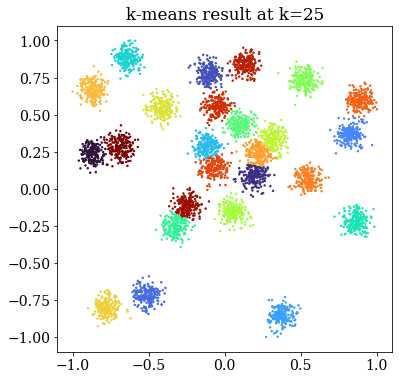}};
\draw[densely dotted,rotate around={-25:(.3,.15)}] (.3,.15) ellipse (.1 and .25);
\end{tikzpicture}
\caption{$k=25$}
\label{fig:many-k25}
\end{subfigure}%
\caption{Clustering results on the ``many blobs'' data set.}
\label{fig:many}
\end{figure*}

We can now generate a standard deviation reduction plot, comparing the
observed with the estimated values:
\begin{align}
\frac{\sqrt{\SSE/(N-k)}}{\sqrt{\widehat{\SSE}_k/(N-k)}}
=
\sqrt{\frac{\SSE}{\widehat{\SSE}_k}}
\end{align}
Note that because $\smash{\widehat{\SSE}_k}$ will tend to~0 as we increase~$k$,
eventually this will become unstable for too large~$k$. 
Depending on our objective, either the smallest value or the last value below~1
(or below a suitable threshold such as 0.99)
can be chosen as ``best'' $k$.
Figure~\ref{fig:reduction} plots this score for some of the above data sets.
The normal distribution never scores below 1, and the uniform distribution remains very close to 1;
hence these data sets can be recognized as unclustered.
For the many blobs data set,
the largest reduction is obtained for $k=8$, and the last reduction is at $k=25$,
the number of generated clusters in this data set. To better understand why both
of these solutions are of interest, Figure~\ref{fig:many} shows that
when going from $k=7$ to $k=8$ we observe structural changes that substantially improved
the clustering result (such as separating the far cluster in the bottom, but also
improving the clustering in the center), whereas the last improvement from $k=24$
to $k=25$ only affected three overlapping blobs in the center, that previously
were split into two and now correctly into three clusters.
For $k=8$ we have the best structural improvement,
but for $k=25$ we get the finest clustering; both may have their use cases.
The solution $k=21$ is preferred by many distance-based measures, it is worth noting
that well-separated blobs are separated, but touching blobs are still joined at this~$k$.

Instead of using the Elbow heuristic, most ``intrinsic'' cluster evaluation
criteria can be used. An extensive survey was published by
Arbelaitz et al.~\cite{DBLP:journals/pr/ArbelaitzGMPP13}, we only discuss a few
examples of particular interest here. This involves some of the best-performing
indexes in this study, namely the Silhouette, the VRC, and the DB-Index.

\pagebreak
\subsection{Variance-based criteria}
So have we found a new, better method to choose the number of clusters~$k$?
The above ``novel'' approach is very similar to the
\textbf{Variance Ratio Criterion (VRC)} published already in the
mid-70s by Calinski and Harabasz~\cite{doi:10.1080/03610927408827101}:
\begin{align}
\operatorname{VRC} :=
\frac{\SSE_1-\SSE_k}{k-1} \Big/ \frac{\SSE_k}{n-k}
\label{vrc}
\end{align}
(using the fact that $\text{BGSS}=\text{TSS}-\text{WGSS}=\SSE_1-\SSE_k$).
As they noted, this is analogous to the F-statistic used when testing
for the significance of a difference in means, but we must not use such
a significance test here, because we optimized the means (which makes
such a test on the difference in means invalid).
There are many more methods discussed in the 70s and 80s literature
overlooked by many machine learning scientists of today,
that we do not have the space to discuss here, we only briefly highlight
some starting points.

Marriott~\cite{journals/biometrics/Marriott71} (and before,
Friedman and Rubin~\cite{journals/jasa/FriedmanR67}) analyze the determinant
of the variance-covariance matrix $|W|$, containing the within-class scatter,
also known as generalized variance, instead of the regular variance
$\operatorname{tr}(W)$, and discuss that the expected change when partitioning
into $k$ clusters is a reduction by $1/k^2$. They argue this approach is
superior because it takes correlations into account.
Krzanowski and Lai~\cite{journal/biometric/KrzanowskiL88} depart from Marriott
and return to using the trace again. They argue that the variance is expected
to decrease by $k^{\frac2d}$, define the successive difference as
$\text{Diff}_k=(k-1)^{\frac2p}\SSE_{k-1}-k^{\frac2p}\SSE_k$, and then find
the maximum of $KL(k):=|\text{Diff}_k/\text{Diff}_{k+1}|$.
When $\text{Diff}_{k+1}$ becomes small, this can become unstable, explaining the poor performance
in our experiments.

Pham et al.~\cite{PhamDN05} propose a scoring function for $k\geq 2$ based
on $\SSE_k/(\alpha_k \SSE_{k-1})$,
where the weights $\alpha_2=1-\frac3{4d}$ and $\alpha_k=\frac56\alpha_{k-1}+\frac16$
model an expected change on a uniform distribution.

\subsection{Distance-based criteria}

The Dunn~\cite{journal/cybernetics/Dunn73} index compares the diameter of clusters
to the cluster separation. It exists in several variations, in the most basic form
it is defined as the ratio of the smallest cluster separation to the largest cluster diameter,
$$
\text{Dunn} := \frac{\min_i \min_{j\neq i} \min_{x\in C_i} \min_{y\in C_y} d(x,y)}%
{\max_i \max_{x\in C_i} \max_{y\in C_i} d(x,y)}
\;.
$$
In this basic (original) version, only the smallest cross-cluster distance and
the largest inter-cluster distance are taken into account, but we might also consider
averages instead~\cite{DBLP:journals/pr/ArbelaitzGMPP13}.
The Davies-Bouldin-Index \cite{journals/tpami/DaviesB79} compares the distance
to the nearest other cluster with the radius of the two clusters.
This is then averaged over all clusters:
$$
\text{DB} := \frac1k \sum\nolimits_i \max_{j\neq i} \frac{S_i+S_j}{M_{ij}}
$$
where $S_i$ is the (arithmetic, or root-mean-square) average distance of points to their cluster center (and hence a kind of radius),
and $M_{ij}$ is the distance between the cluster centers.
When using the root-mean-square averages, $S_i$ is the average distance of points within the cluster,
and $M_{ij}$ is the average distance between points in different clusters.

One of the most used distance-based criteria is the
average silhouette width measure~\cite{Rousseeuw/87a},
which compares the average distance of each point to its own cluster to the average
distance to the nearest other cluster. This method is closely related to
$k$-medoids clustering and the PAM algorithm~\cite{Kaufman/Rousseeuw/87a,DBLP:journals/is/SchubertR21},
which cluster the data around $k$ representative objects (called medoids),
minimizing the distances to the medoids. In contrast to $k$-means (which minimizes
squared errors), this method can also be used to optimize Euclidean or Manhattan
distance; but it is mostly of interest for distances where the mean is not useful.%
\footnote{While the arithmetic means in $k$-means do \emph{not} minimize Euclidean
or Manhattan distance, it is often good enough to be useful for many applications.} 
As Silhouette is fairly expensive to compute, it can be simplified by using the
distance from the cluster center or medoid instead of the average distance.
But it turns out that we can try to directly optimize this measure using
PAM-like algorithms \cite{VanderLaan/03a,DBLP:conf/sisap/LenssenS22}.

\subsection{Information-theoretic criteria}
A different idea to choose the optimum number of clusters is based on the
principle of minimum description length. Here, a $k$-means solution is considered
better, if the data can be encoded more compactly. Increasing the number of
centers means that the input data is approximated more closely (and hence needs
less to encode the deviations), but at the same time, we also need to store more
cluster centers. This intuition nicely fits the idea of approximating data
and data quantization.
X-means~\cite{DBLP:conf/icml/PellegM00} integrates this with $k$-means in an
algorithm that dynamically increases the number of clusters as long as a cluster
quality criterion improves. They proposed to use the Bayesian Information Criterion~(BIC)
of Schwarz~\cite{10.1214/aos/1176344136}, who also proposed the Akaike Information Criterion~(AIC).
The original X-means version appears to have an error,
the fixed equation of Foglia and Hancock~\cite{web/FogliaH12} appears to work better.
G-means~\cite{DBLP:conf/nips/HamerlyE03} uses Anderson-Darling tests instead to decide
when to accept a new cluster, and when to stop increasing~$k$.

\subsection{Simulation-based criteria}
Tibshirani et al.~\cite{journals/jrss/TibshiraniWH01} propose the gap statistic,
which estimates a baseline $\SSE^\prime_k$ obtained by clustering uniform random data sets.
They then choose a $k$ using $$\text{Gap}_k:=E[\log \SSE_k^\prime] - \log\SSE_k\;,$$ and picking
the smallest $k$ such that $\text{Gap}_k{\geq} \text{Gap}_{k-1} {-} s_{k+1}$ where $s_{k+1}$
is the standard deviation of the estimates.
This works decently well for synthetic data, but most interestingly it failed to recognize the uniform
data as unclustered. For the more challenging data sets, the estimated number of clusters
was unstable with the default sample sizes.

As we are not convinced that the ``novel'' approach we constructed above is clearly
superior to VRC, BIC, or the Gap statistic,
we suggest that you simply use one of these approaches to choose~$k$,
and rather pay attention to the way you preprocess your data for $k$-means.
because ``garbage in, garbage out'' -- if your data is not prepared well,
none of the clustering results will be good.

\section{The true challenges of k-means}

\begin{figure*}[t]
\begin{subfigure}[t]{.19\linewidth}\centering\vskip0pt
\includegraphics[height=\linewidth]{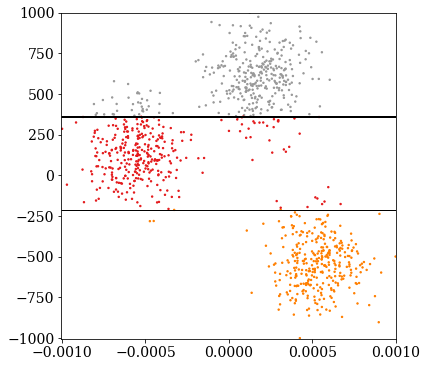}
\caption{Poorly scaled}
\label{fig:challenges-scaled}
\end{subfigure}%
\hfill%
\begin{subfigure}[t]{.19\linewidth}\centering\vskip0pt
\includegraphics[height=\linewidth]{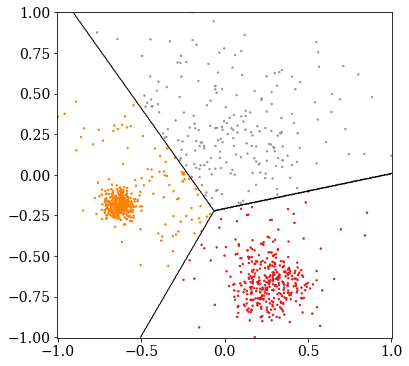}
\caption{Varying diameter}
\label{fig:challenges-diameter}
\end{subfigure}%
\hfill%
\begin{subfigure}[t]{.19\linewidth}\centering\vskip0pt
\includegraphics[height=\linewidth]{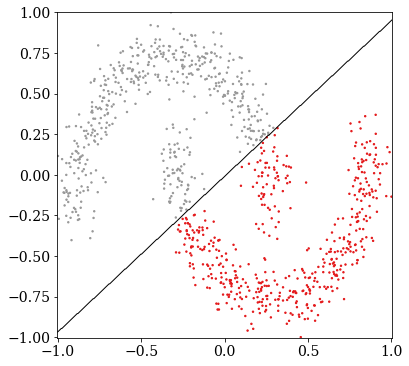}
\caption{Non-convex}
\label{fig:challenges-nonconvex}
\end{subfigure}%
\hfill%
\begin{subfigure}[t]{.19\linewidth}\centering\vskip0pt
\includegraphics[height=\linewidth]{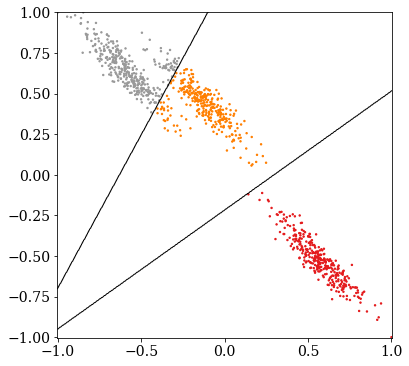}
\caption{Correlated}
\label{fig:challenges-correlated}
\end{subfigure}%
\hfill%
\begin{subfigure}[t]{.19\linewidth}\centering\vskip0pt
\includegraphics[height=\linewidth]{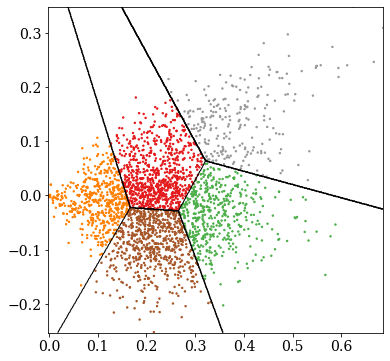}
\caption{Poorly transformed}
\label{fig:challenges-transformed}
\end{subfigure}%
\caption{Examples of data sets where $k$-means cannot be expected to work well.}
\label{fig:challenges}
\end{figure*}
While the difficulty of choosing $k$ is easily noticed by the user,
as he has to specify this parameter, it nevertheless remains much more
difficult to obtain meaningful and useful results from $k$-means than
commonly anticipated.
If we study the foundations of $k$-means, and the relationship to Gaussian
mixture modeling, we can observe that $k$-means assumes errors
to be invariant across the entire data space, whereas in full Gaussian mixture modeling,
the deviation from a cluster in certain directions weight less than in other,
and depend on the individual clusters.
We can consider $k$-means as a limit case of Gaussian mixture modeling, where we
perform (i)~all clusters have an identical, spherical shape,
and (ii)~we make hard cluster assignments, for example by making the cluster standard deviations tend to zero.
We will not go into (ii) here in detail (see, e.g., Bishop~\cite{DBLP:books/lib/Bishop07}).
The observation of interest is that in $k$-means we somewhat assume that all clusters have the same spherical shape.
This is simply a consequence of the sum of squared errors (Eq.~\ref{eq:sse})
not including any weights depending on the cluster or axis.
This is a reasonable simplification if we assume that our data set was generated
from $k$ pure signals (corresponding to the cluster centers)
plus i.i.d.{} Gaussian noise.

This also leads to many situations where $k$-means will not work well:
for example (i)~if the axes have very different scales, and clusters are separated
on the scales of low variance as in Figure~\ref{fig:challenges-scaled},
(ii)~if the cluster diameters are very different, yet the clusters are close,
as in Figure~\ref{fig:challenges-diameter}
and (iii)~when the clusters are not generated by Gaussian errors around an origin
but have a non-convex shape as in Figure~\ref{fig:challenges-nonconvex} and common in geodata,
(iv)~there are correlations in the data and some directions are more important
than others, as in Figure~\ref{fig:challenges-correlated},
(v)~the input data is not continuous,
or (vi)~the similarity of objects is not well captured by Euclidean distance.
When dealing with complex data, such as text data, it is fairly common that we first have to
``vectorize'' it, for example using TF-IDF, and/or applying some dimensionality reduction technique
such as principal components analysis (PCA).
Figure~\ref{fig:challenges-transformed} shows such a data set, containing 5 groups from the well-known
20newsgroups data set, reduced to two dimensions with TF-IDF and PCA. It exhibits a typical ``conical''
shape with a tip at the zero, then extending into the first component, often seen with PCA on sparse input data,
with the first component often capturing an overall vector length. In this case, all signal suitable for
clustering was destroyed by this na\"ive preprocessing -- yet this is a common combination of preprocessing techniques
recommended in various blogs.
The first two cases and the fourth case can be solved much better
by using Gaussian mixture modeling~\cite{DempsterLR77}, in the third case -- common for example in data
that follows geographic features -- DBSCAN~\cite{DBLP:conf/kdd/EsterKSX96,DBLP:journals/tods/SchubertSEKX17} and other
density-based clustering algorithms often are a better choice.
For (v), the choice of a suitable clustering algorithm tends to become difficult,
although modifications of $k$-means to categorical variables exist,
for example, the $k$-modes algorithm~\cite{DBLP:journals/datamine/Huang98};
There exist many clustering algorithms that allow using other distance functions,
such as classic hierarchical clustering,
$k$-medoids clustering~\cite{KaufmanR90,DBLP:conf/sisap/SchubertR19},
spherical $k$-means~\cite{DBLP:journals/ml/DhillonM01,DBLP:conf/sisap/SchubertLF21}
and DBSCAN~\cite{DBLP:conf/kdd/EsterKSX96,DBLP:journals/tods/SchubertSEKX17}.
When looking at clustering results reported using $k$-means, in many cases the results suffer from
at least one of these additional problems as well.

\section{Conclusion}
Given the prevalence of the elbow method in education,
online media (such as Wikipedia\footnote{E.g, \url{https://en.wikipedia.org/w/index.php?title=Elbow_method_(clustering)&oldid=1099441401} as of 2022}),
and even clustering research (as evidenced by the many proposals to automatically identify an elbow),
it appears to be due to warn of using this method and to emphasize that
much better alternatives such as the variance-ratio criterion (VRC) of
Calinski and Harabasz \cite{doi:10.1080/03610927408827101},
the Bayesian Information Criterion (BIC),
or the Gap statistics should always be preferred instead.
While the problems of the elbow approach have been discussed several times in the literature
(e.g., \cite{Milligan1985,KetchenS96}),
this knowledge of clustering basics appears to have been largely forgotten in
today's machine learning community and hence needs to be communicated again.
Educators should omit the method or at least explain better alternatives. 
Data scientists must be made wary of drawing conclusions
from clustering results because of such problems, and
must not rely on evaluation measures telling them what is ``best''.
Reviewers of scientific literature should probably even reject conclusions drawn from
choosing the ``optimal'' $k$ using such an unreliable method. In the long run,
we must accept that there is no ``optimal'' solution in cluster analysis, but it is
an explorative approach that may yield multiple interesting solutions, and
interestingness necessarily is a subjective decision of the user.

\pagebreak
\bibliography{literature} 

\begin{thebibliography}{10}

\bibitem{DBLP:journals/ml/AloiseDHP09}
{\sc Aloise, D., Deshpande, A., Hansen, P., and Popat, P.}
\newblock {NP}-hardness of euclidean sum-of-squares clustering.
\newblock {\em Mach. Learn. 75}, 2 (2009), 245--248.

\bibitem{DBLP:journals/pr/ArbelaitzGMPP13}
{\sc Arbelaitz, O., Gurrutxaga, I., Muguerza, J., P{\'{e}}rez, J.~M., and
  Perona, I.}
\newblock An extensive comparative study of cluster validity indices.
\newblock {\em Pattern Recognit. 46}, 1 (2013), 243--256.

\bibitem{DBLP:books/lib/Bishop07}
{\sc Bishop, C.~M.}
\newblock {\em Pattern recognition and machine learning, 5th Edition}.
\newblock Information science and statistics. Springer, 2007.

\bibitem{Bock2007}
{\sc Bock, H.-H.}
\newblock {\em Clustering Methods: A History of k-Means Algorithms}.
\newblock Springer, 2007, pp.~161--172.

\bibitem{DBLP:journals/ibmrd/Bonner64}
{\sc Bonner, R.~E.}
\newblock On some clustering techniques.
\newblock {\em {IBM} J. Res. Dev. 8}, 1 (1964), 22--32.

\bibitem{doi:10.1080/03610927408827101}
{\sc Caliński, T., and Harabasz, J.}
\newblock A dendrite method for cluster analysis.
\newblock {\em Communications in Statistics 3}, 1 (1974), 1--27.

\bibitem{journals/tpami/DaviesB79}
{\sc Davies, D.~L., and Bouldin, D.~W.}
\newblock A cluster separation measure.
\newblock {\em IEEE Transactions on Pattern Analysis and Machine Intelligence
  PAMI-1}, 2 (1979), 224--227.

\bibitem{DempsterLR77}
{\sc Dempster, A.~P., Laird, N.~M., and Rubin, D.~B.}
\newblock Maximum likelihood from incomplete data via the em algorithm.
\newblock {\em Journal of the Royal Statistical Society: Series B
  (Methodological) 39}, 1 (1977), 1--22.

\bibitem{DBLP:journals/ml/DhillonM01}
{\sc Dhillon, I.~S., and Modha, D.~S.}
\newblock Concept decompositions for large sparse text data using clustering.
\newblock {\em Mach. Learn. 42}, 1/2 (2001), 143--175.

\bibitem{journal/cybernetics/Dunn73}
{\sc Dunn, J.~C.}
\newblock A fuzzy relative of the isodata process and its use in detecting
  compact well-separated clusters.
\newblock {\em Journal of Cybernetics 3}, 3 (1973), 32--57.

\bibitem{DBLP:conf/kdd/EsterKSX96}
{\sc Ester, M., Kriegel, H., Sander, J., and Xu, X.}
\newblock A density-based algorithm for discovering clusters in large spatial
  databases with noise.
\newblock In {\em Knowledge Discovery and Data Mining, {KDD}\/} (1996),
  pp.~226--231.

\bibitem{DBLP:journals/sigkdd/Estivill-Castro02}
{\sc Estivill{-}Castro, V.}
\newblock Why so many clustering algorithms: a position paper.
\newblock {\em {SIGKDD} Explor. 4}, 1 (2002), 65--75.

\bibitem{web/FogliaH12}
{\sc Foglia, A., and Hancock, B.}
\newblock Notes on bayesian information criterion calculation for x-means
  clustering, 2012.

\bibitem{journals/jasa/FriedmanR67}
{\sc Friedman, H.~P., and Rubin, J.}
\newblock On some invariant criteria for grouping data.
\newblock {\em Journal of the American Statistical Association 62}, 320 (1967),
  1159--1178.

\bibitem{DBLP:conf/nips/HamerlyE03}
{\sc Hamerly, G., and Elkan, C.}
\newblock Learning the k in k-means.
\newblock In {\em Neural Information Processing Systems, {NIPS}\/} (2003),
  pp.~281--288.

\bibitem{DBLP:journals/datamine/Huang98}
{\sc Huang, Z.}
\newblock Extensions to the k-means algorithm for clustering large data sets
  with categorical values.
\newblock {\em Data Min. Knowl. Discov. 2}, 3 (1998), 283--304.

\bibitem{Kaufman/Rousseeuw/87a}
{\sc Kaufman, L., and Rousseeuw, P.~J.}
\newblock Clustering by means of medoids.
\newblock In {\em Statistical Data Analysis Based on the $L_1$ Norm and Related
  Methods}, Y.~Dodge, Ed. North-Holland, 1987, pp.~405--416.

\bibitem{KaufmanR90}
{\sc Kaufman, L., and Rousseeuw, P.~J.}
\newblock Partitioning around medoids (program {PAM}).
\newblock In {\em Finding Groups in Data: An Introduction to Cluster Analysis}.
  John Wiley \& Sons, Ltd, 1990, ch.~2, pp.~68--125.

\bibitem{KetchenS96}
{\sc Ketchen, D.~J., and Shook, C.~L.}
\newblock The application of cluster analysis in strategic management research:
  An analysis and critique.
\newblock {\em Strategic Management Journal 17}, 6 (1996), 441--458.

\bibitem{journal/biometric/KrzanowskiL88}
{\sc Krzanowski, W.~J., and Lai, Y.~T.}
\newblock A criterion for determining the number of groups in a data set using
  sum-of-squares clustering.
\newblock {\em Biometrics 44}, 1 (1988), 23--34.

\bibitem{DBLP:conf/sisap/LenssenS22}
{\sc Lenssen, L., and Schubert, E.}
\newblock Clustering by direct optimization of the medoid silhouette.
\newblock In {\em Similarity Search and Applications\/} (2022), pp.~190--204.

\bibitem{DBLP:conf/walcom/MahajanNV09}
{\sc Mahajan, M., Nimbhorkar, P., and Varadarajan, K.~R.}
\newblock The planar k-means problem is {NP}-hard.
\newblock In {\em {WALCOM:} Algorithms and Computation\/} (2009), pp.~274--285.

\bibitem{journals/biometrics/Marriott71}
{\sc Marriott, F. H.~C.}
\newblock Practical problems in a method of cluster analysis.
\newblock {\em Biometrics 27}, 3 (1971), 501--514.

\bibitem{Milligan1985}
{\sc Milligan, G.~W., and Cooper, M.~C.}
\newblock An examination of procedures for determining the number of clusters
  in a data set.
\newblock {\em Psychometrika 50}, 2 (June 1985), 159--179.

\bibitem{Novikov2019}
{\sc Novikov, A.}
\newblock {PyClustering}: Data mining library.
\newblock {\em Journal of Open Source Software 4}, 36 (2019), 1230.

\bibitem{Onumanyi22}
{\sc Onumanyi, A.~J., Molokomme, D.~N., Isaac, S.~J., and Abu-Mahfouz, A.~M.}
\newblock Autoelbow: An automatic elbow detection method for estimating the
  number of clusters in a dataset.
\newblock {\em Applied Sciences 12}, 15 (2022).

\bibitem{DBLP:conf/icml/PellegM00}
{\sc Pelleg, D., and Moore, A.~W.}
\newblock X-means: Extending k-means with efficient estimation of the number of
  clusters.
\newblock In {\em Int. Conf. Machine Learning {(ICML})\/} (2000), pp.~727--734.

\bibitem{PhamDN05}
{\sc Pham, D.~T., Dimov, S.~S., and Nguyen, C.~D.}
\newblock Selection of k in k-means clustering.
\newblock {\em Proceedings of the Institution of Mechanical Engineers, Part C:
  Journal of Mechanical Engineering Science 219}, 1 (2005), 103--119.

\bibitem{Rousseeuw/87a}
{\sc Rousseeuw, P.~J.}
\newblock Silhouettes: A graphical aid to the interpretation and validation of
  cluster analysis.
\newblock {\em J. Comput. Appl. Math. 20\/} (1987), 53--65.

\bibitem{DBLP:conf/ictai/SalvadorC04}
{\sc Salvador, S., and Chan, P.}
\newblock Determining the number of clusters/segments in hierarchical
  clustering/segmentation algorithms.
\newblock In {\em Tools with Artificial Intelligence {(ICTAI)}\/} (2004),
  pp.~576--584.

\bibitem{DBLP:conf/icdcsw/SatopaaAIR11}
{\sc Satopää, V., Albrecht, J.~R., Irwin, D.~E., and Raghavan, B.}
\newblock Finding a "kneedle" in a haystack: Detecting knee points in system
  behavior.
\newblock In {\em Distributed Computing Systems ({ICDCS}) Workshops\/} (2011),
  pp.~166--171.

\bibitem{DBLP:conf/sisap/SchubertLF21}
{\sc Schubert, E., Lang, A., and Feher, G.}
\newblock Accelerating spherical k-means.
\newblock In {\em Similarity Search and Applications\/} (2021), pp.~217--231.

\bibitem{DBLP:conf/sisap/SchubertR19}
{\sc Schubert, E., and Rousseeuw, P.~J.}
\newblock Faster k-medoids clustering: Improving the {PAM}, {CLARA}, and
  {CLARANS} algorithms.
\newblock In {\em Similarity Search and Applications, {SISAP}\/} (2019),
  pp.~171--187.

\bibitem{DBLP:journals/is/SchubertR21}
{\sc Schubert, E., and Rousseeuw, P.~J.}
\newblock Fast and eager k-medoids clustering: {O}(k) runtime improvement of
  the {PAM}, {CLARA}, and {CLARANS} algorithms.
\newblock {\em Inf. Syst. 101\/} (2021), 101804.

\bibitem{DBLP:journals/tods/SchubertSEKX17}
{\sc Schubert, E., Sander, J., Ester, M., Kriegel, H., and Xu, X.}
\newblock {DBSCAN} revisited, revisited: Why and how you should (still) use
  {DBSCAN}.
\newblock {\em {ACM} Trans. Database Syst. 42}, 3 (2017), 19:1--19:21.

\bibitem{10.1214/aos/1176344136}
{\sc Schwarz, G.}
\newblock {Estimating the Dimension of a Model}.
\newblock {\em The Annals of Statistics 6}, 2 (1978), 461 -- 464.

\bibitem{DBLP:journals/ejwcn/ShiWWWLL21}
{\sc Shi, C., Wei, B., Wei, S., Wang, W., Liu, H., and Liu, J.}
\newblock A quantitative discriminant method of elbow point for the optimal
  number of clusters in clustering algorithm.
\newblock {\em {EURASIP} J. Wirel. Commun. Netw. 2021}, 1 (2021), 31.

\bibitem{journals/jasa/SugarJ03}
{\sc Sugar, C.~A., and James, G.~M.}
\newblock Finding the number of clusters in a dataset: An information-theoretic
  approach.
\newblock {\em Journal of the American Statistical Association 98}, 463 (2003),
  750--763.

\bibitem{Thorndike53}
{\sc Thorndike, R.~L.}
\newblock Who belongs in the family?
\newblock {\em Psychometrika 18}, 4 (1953), 267--276.

\bibitem{journals/jrss/TibshiraniWH01}
{\sc Tibshirani, R., Walther, G., and Hastie, T.}
\newblock Estimating the number of clusters in a data set via the gap
  statistic.
\newblock {\em Journal of the Royal Statistical Society: Series B (Statistical
  Methodology) 63}, 2 (2001), 411--423.

\bibitem{VanderLaan/03a}
{\sc {Van der Laan}, M., Pollard, K., and Bryan, J.}
\newblock A new partitioning around medoids algorithm.
\newblock {\em Journal of Statistical Computation and Simulation 73}, 8 (2003),
  575--584.

\bibitem{DBLP:journals/isci/ZhangMQG17}
{\sc Zhang, Y., Mandziuk, J., Quek, H.~C., and Goh, W.}
\newblock Curvature-based method for determining the number of clusters.
\newblock {\em Inf. Sci. 415\/} (2017), 414--428.

\end{thebibliography}
\bibliographystyle{acm}

\end{document}